\pdfoutput=1
\documentclass[11pt]{article} 

\usepackage[final]{naacl2021}

\usepackage{times}
\usepackage{latexsym}
\usepackage{xspace}
\usepackage{tabularx, ragged2e, caption, lipsum}

\usepackage[T1]{fontenc}
\usepackage[utf8]{inputenc}

\usepackage{float}
\usepackage{multicol}
\usepackage{multirow}
\usepackage{graphicx}
\usepackage{makecell}

\usepackage{microtype}

\newcommand{\dataset}{\textsc{clinsum}\xspace}
\newcommand{\xhdr}[1]{\noindent{{\bf #1.}}}

\title{What's in a Summary? \\ Laying the Groundwork for Advances in Hospital-Course Summarization}

\author{Griffin Adams$^{1}$, Emily Alsentzer$^{2}$, Mert Ketenci$^{1}$, Jason Zucker$^{1}$, No\'emie Elhadad$^{1}$ \\
  $^{1}$Columbia University, New York, NY \\
  $^{2}$Harvard-MIT's Health Science and Technology, Cambridge, MA \\
  \texttt{\{griffin.adams, mert.ketenci, noemie.elhadad\}@columbia.edu} \\
  \texttt{emilya@mit.edu}, \texttt{jz2700@cumc.columbia.edu}
}

\begin{document}
\maketitle
\begin{abstract}

Summarization of clinical narratives is a long-standing research problem. Here, we introduce the task of hospital-course summarization. Given the documentation authored throughout a patient's hospitalization, generate a paragraph that tells the story of the patient admission. We construct an English, text-to-text dataset of 109,000 hospitalizations (2M source notes) and their corresponding summary proxy: the clinician-authored ``Brief Hospital Course'' paragraph written as part of a discharge note. Exploratory analyses reveal that the BHC paragraphs are highly abstractive with some long extracted fragments; are concise yet comprehensive; differ in style and content organization from the source notes;  exhibit minimal lexical cohesion; and represent silver-standard references. Our analysis identifies multiple implications for modeling this complex, multi-document summarization task.

\end{abstract}

\section{Introduction}

The electronic health record (EHR) contains critical information for clinicians to assess a patient's medical history (e.g., conditions, laboratory tests, procedures, treatments) and healthcare interactions (e.g., primary care and specialist visits, emergency department visits, and hospitalizations). While medications, labs, and diagnoses are documented through structured data elements and flowsheets, clinical notes contain rich narratives describing the patient's medical condition and interventions. A single hospital visit for a patient with a lengthy hospital stay, or complex illness, can consist of hundreds of notes. At the point of care, clinicians already pressed for time, face a steep challenge of making sense of their patient's documentation and synthesizing it either for their own decision making process or to ensure coordination of care~\citep{hall2004information,ash2004some}.

Automatic summarization has been proposed to support clinicians in multiple scenarios, from making sense of a patient's longitudinal record over long periods of time and multiple interactions with the healthcare system, to synthesizing a specific visit's documentation. Here, we focus on \textit{hospital-course summarization}: faithfully and concisely summarizing the EHR documentation for a patient's specific inpatient visit, from admission to discharge. Crucial for continuity of care and patient safety after discharge~\citep{kripalani2007deficits,van2002effect}, hospital-course summarization also represents an incredibly challenging multi-document summarization task with diverse knowledge requirements.  To properly synthesize an admission, one must not only identify relevant problems, but link them to symptoms, procedures, medications, and observations while adhering to temporal, problem-specific constraints.

% These works incorporate medical ontologies into standard encoder-decoder frameworks by using an entity-aware pointer-generator network~\citep{macavaney2019ontology} or by leveraging attention over a soft-selection of likely to be copied entities~\citep{sotudeh2020attend}. \citet{zhang2019optimizing} directly optimize factualness by training with a factual correctness reward via reinforcement learning.

Our main contributions are as follows: (1) We introduce the task of hospital-course summarization; (2) we collect a dataset of inpatient documentation and corresponding "Brief Hospital Course" paragraphs extracted from discharge notes; and (3) we assess the characteristics of these summary paragraphs as a proxy for target summaries and discuss implications for the design and evaluation of a hospital-course summarization tool.

\section{Related Works}

Summarization of clinical data and documentation has been explored in a variety of use cases \citep{pivovarov2015automated}. For longitudinal records, graphical representations of structured EHR data elements (i.e., diagnosis codes, laboratory test measurements, and medications) have been proposed~\citep{powsner1997summarizing,plaisant1996lifelines}. Interactive visualizations of clinical problems' salience, whether extracted from notes~\citep{hirsch2015harvest} or inferred from clinical documentation~\cite{levy2020towards} have shown promise~\citep{pivovarov2016can,2020levyfixthesis}.

Most work in this area, however, has focused on clinical documentation of a fine temporal resolution. Traditional text generation techniques have been proposed to synthesize structured data like ICU physiological data streams ~\citep{hunter2008summarising,goldstein2016automated}. \citet{liu2018learning} use a transformer model to write EHR notes from the prior 24 hours, while \citet{liang2019novel} perform disease-specific summarization from individual progress notes. 
\citet{mcinerney2020query} develop a distant supervision approach to generate extractive summaries to aid radiologists when interpreting images.
\citet{zhang2018learning, zhang2019optimizing, macavaney2019ontology, sotudeh2020attend} generate the ``Impression'' section of the Radiology report from the more detailed ``Findings'' section.  Finally, several recent works aim to generate EHR notes from doctor-patient conversations~\citep{krishna2020generating,joshi2020dr, MSR_2020}. Recent work on summarizing hospital admissions focuses on extractive methods \citep{moen2014evaluation,moen2016comparison, liu2018unsupervised, alsentzer2018extractive}.

\section{Hospital-Course Summarization Task}

Given the clinical documentation available for a patient hospitalization, our task of interest is to generate a text that synthesizes the hospital course in a faithful and concise fashion.  For our analysis, we rely on the ``Brief Hospital Course'' (BHC), a mandatory section of the discharge note, as a proxy reference.  The BHC tells the story of the patient's admission: \textit{what} was done to the patient during the hospital admission and \textit{why}, as well as the \textit{follow up} steps needed to occur post discharge, whenever needed. Nevertheless, it is recognized as a challenging and time consuming task for clinicians to write~\cite{dodd07,ucirvine20}.
 
\subsection{Dataset}

To carry out our analysis, we construct a large-scale, multi-document summarization dataset, \dataset. Materials come from all hospitalizations between 2010 and 2014 at Columbia University Irving Medical Center. \textbf{Table~\ref{tab:clinsum-stats}} shows summary statistics for the corpus. There are a wide range of reasons for hospitalizations, from life-threatening situations (e.g., heart attack) to when management of a specific problem cannot be carried out effectively outside of the hospital (e.g., uncontrolled diabetes). This contributes to the high variance in documentation. For reference, Table~\ref{tab:corpora} provides a comparison of basic statistics to widely used summarization datasets.  Relatively speaking, \dataset is remarkable for having a very high compression ratio despite having long reference summaries.  Additionally, it appears highly extractive with respect to fragment density (we qualify this in Section \ref{sec:extractiveness}).

Based on advice from clinicians, we rely on the following subset of note types as source documents: ``Admission'', ``Progress'', and ``Consult'' notes. The dataset does not contain any structured data, documentation from past encounters, or other note types (e.g., nursing notes, social work, radiology reports)~\citep{reichert2010cognitive}.  Please refer to Appendix \ref{app:sum-datasets} for more details and rationale.

\begin{table}[t]
    \small
    \centering
    \begin{tabular}{c|l|c|c}
    & \bfseries Variable & \bfseries Value & \bfseries STD \\
    % \hline
    \multirow{3}{*}{Global}
     & \# Patients & 68,936 & \multirow{3}{*}{N/A} \\
     & \# Admissions & 109,726 &  \\
     & \# Source Notes & 2,054,828 &  \\ 
    \hline
    \multirow{5}{*}{\makecell{Per \\ Adm.}}
    & Length of Stay & 5.8 days & 9.0 \\
    & \# Source Notes & 18.7 & 30.1 \\
    & \# Source Sentences & 1,061.2 & 1,853.6 \\
    & \# Source Tokens & 11,838.7 & 21,506.5 \\
    & \# Summary Sentences & 17.8 & 16.9 \\
    & \# Summary Tokens & 261.9 & 233.8 \\
    \hline
    \multirow{2}{*}{\makecell{Per \\ Sent.}}
    & \# Source Tokens & 10.9 & 12.4 \\
    & \# Summary Tokens & 14.5 & 11.5 \\
    \hline
    \multirow{1}{*}{Ratio}
    % & Sentence Compression & 59.6 & TBD \\
    & Word Compression & 42.5 & 164.6 \\
    \end{tabular}
    \caption{Basic Statistics for \dataset.  Value is the total for Global, and average for `Per Admission' and `Per Sentence'. STD is standard deviation.}
    \label{tab:clinsum-stats}
\end{table}

\subsection{Tools for Analysis}

\paragraph{Entity Extraction \& Linking.} 
We use the MedCAT toolkit \citep{kraljevic2020multi} to extract medical entity mentions and normalize to concepts from the UMLS (Unified Medical Language System) terminology~\citep{bodenreider2004unified}.  To exclude less relevant entities, we only keep entities from the Disorders, Chemicals \& Drugs, and Procedures semantic groups, or the Lab Results semantic type.

\paragraph{Local Coherence.} We examine inter-sentential coherence in two ways. \textbf{Next-Sentence Prediction (NSP)}. Since we compare across a few datasets representing different domains, we use domain-specific pre-trained BERT models via HuggingFace \citep{wolf2019huggingface}: ``bert-base-cased'' for CNN/DM and Arxiv, ``monologg/biobert\_v1.1\_pubmed'' for Pubmed, and ``emilyalsentzer/Bio\_ClinicalBERT'' for \dataset. \textbf{Entity-grids.} Entity-grids model local coherence by considering the distribution of discourse entities \citep{barzilay2008modeling}. An entity grid is a 2-D representation of a text whose entries represent the presence or absence of a discourse entity in a sentence.  For our analyses, we treat UMLS concepts as entities and train a neural model, similar to \citet{nguyen2017neural, mohiuddin2018coherence}, which learns to rank the entity grid of a text more highly than the same entity grid whose rows (sentences) have been randomly shuffled.  Please see Appendix \ref{app:coherence} for more details.

\paragraph{Lexical Overlap Metric.} We use ROUGE-1 (R1) \& ROUGE-2 (R2) F-1~\citep{lin2004rouge} to measure lexical overlap, while ignoring higher order variants based on analysis from other work \citep{krishna2021}.  We denote the average of R1 \& R2 scores as $R_{12}$.

\paragraph{Extractive Summarization Baselines.} We rely on a diverse set of sentence extraction methods, whose performance on a held-out portion of \dataset is reported in Table \ref{tab:big-table}. \textbf{Oracle models} have access to the ground-truth reference and represent upper bounds for extraction.  Here, we define the sentence selection criteria for each oracle variant, leaving more in-depth discussion to the subsequent analysis. \textbf{\textsc{Oracle Top-K}}: Take sentences with highest $R_{12}$ vis-a-vis the reference until a target token count is reached; \textbf{\textsc{Oracle Gain}}: Greedily take source sentence with highest relative $R_{12}$ gain conditioned on existing summary\footnote{This is the Neusum model's objective \citep{zhou2018neural}}. Extract sentences until the change in $R_{12}$ is negative;
\textbf{\textsc{Oracle Sent-Align}}: For each sentence in reference, take source sentence with highest $R_{12}$ score;
\textbf{\textsc{Oracle Retrieval}}: For each sentence in reference, take reference sentence from train set with largest BM25 score \citep{robertson1994some}; and 
\textbf{\textsc{Oracle Sent-Align + Retrieval}}: For each sentence in reference, take sentence with highest $R_{12}$ between \textsc{Oracle Sent-Align} and \textsc{Oracle Retrieval}.  We provide two \textbf{unsupervised methods} as well. \textbf{\textsc{Random}}: extracts random sentences until summary reaches target word count (average summary length); \textbf{\textsc{LexRank}}: selects the top-k sentences with largest LexRank \citep{erkan2004lexrank} score until target word count is reached.  For a supervised baseline, we present \textbf{\textsc{ClinNeusum}}: a variant of the Neusum model adapted to the clinical genre~\citep{zhou2018neural}.  \textsc{ClinNeusum} is a hierarchical LSTM network trained on ground-truth labels derived from \textsc{Oracle Gain}, which we detail in Appendix \ref{clin-neusum-details}.

\begin{table*}[ht]
    \centering
    \begin{tabular}{l|ccc|ccc}
    \bfseries Extractive Baseline & \multicolumn{3}{c}{\textbf{ROUGE-1}} & \multicolumn{3}{c}{\textbf{ROUGE-2}} \\
    \hline
     & Recall & Precision & F1 & Recall & Precision & F1  \\
     \hline
    \textsc{Random} & 0.16 & 0.24 & 0.17 & 0.04 & 0.03 & 0.03 \\
    \textsc{LexRank} & 0.18 & 0.21 & 0.18 & 0.05 & 0.05 & 0.05 \\
    \textsc{ClinNeusum} & 0.36 & 0.25 & 0.27 & 0.14 & 0.1 & 0.11 \\
    \textsc{Oracle Top-K} & 0.28 & 0.52 & 0.32 & 0.16 & 0.32 & 0.19 \\
    \textsc{Oracle Gain} & 0.43 & 0.63 & 0.5 & 0.26 & 0.42 & 0.3 \\
    \textsc{Oracle Sent-Align (SA)} & 0.48 & 0.61 & 0.52 & 0.3 & 0.33 & 0.31 \\
    \textsc{Oracle Retrieval} & 0.51 & 0.70 & 0.58 & 0.25 & 0.28 & 0.29 \\
    \textsc{Oracle SA + Retrieval} & \textbf{0.6} & \textbf{0.76} & \textbf{0.66} & \textbf{0.4} & \textbf{0.49} & \textbf{0.43} \\
    \end{tabular}
    \caption{Performance of different sentence selection strategies on \dataset.}
    \label{tab:big-table}
    \vskip -0.1in
\end{table*}

\section{Dataset Analysis \& Implications}

To motivate future research in multiple, self-contained directions, we distill task-specific characteristics to a few salient, standalone takeaways.  For each takeaway, we provide evidence in the data and/or literature, before proposing implications of findings on model development and evaluation. 

% We connect recent developments in summarization research to our task, as well as put forth task-specific novel variants when appropriate.

\subsection{Summaries are mostly abstractive with a few long segments of copy-pasted text}
\label{sec:extractiveness}

\paragraph{tl;dr.} \dataset summaries appear extractive according to widely used metrics.  Yet, there is large variance within summaries.  This directly affects the performance of a supervised extractive model, whose selection capability degrades as summary content transitions from copy-paste to abstractive.  In turn, we need models which can handle abrupt transitions between extractive and abstractive text.

\paragraph{Background.} Clinicians copy forward information from previous notes to save time and ensure that each note includes sufficient evidence for billing and insurance purposes \citep{wrenn2010quantifying}.  Copy-paste is both widely used (66-90\% of clinicians according to a recent literature review \citep{tsou2017safe}) and widely applied (a recent study concluded that in a typical note, 18\% of the text was manually entered; 46\%, copied; and 36\% imported\footnote{Imported refers to text typically pulled in from structured data, such as a medication or problem list.}
\citep{wang2017characterizing}). Please see Appendix \ref{app:note-on-copypaste} for more information on the issue of copy-paste.

\paragraph{Analysis - extractiveness.}

\dataset appears very extractive: a high coverage (0.83 avg / 0.13 std) and a very high density (13.1 avg / 38.0 std) (See \citet{grusky2018newsroom} for a description of the statistics).  However, we find that 64\% of the extractive fragments are unigrams, and 25\% are bigrams, which indicate a high level of re-writing.  The density measure is large because the remaining 11\% of extractive fragments are very long.

\begin{figure}[ht]
\begin{center}
\centerline{\includegraphics[width=\columnwidth]{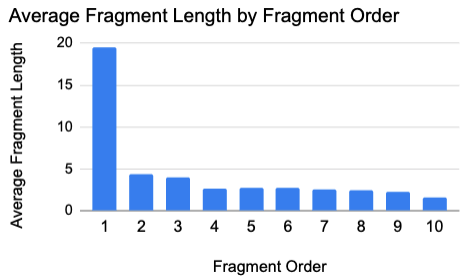}}
\caption{Average extractive fragment lengths according to their relative order within the summary.}
\label{fig:frag-len-rank}
\end{center}
\vskip -0.2in
\end{figure}

Yet, there is a strong positional bias within summaries for long fragments. \textbf{Figure \ref{fig:frag-len-rank}}, groups fragments according to their relative order within each summary.  The longest fragments are usually first.  Qualitative analysis confirms that the beginning of the BHC is typically copied from a previous note and conveys the ``one-liner'' (e.g., \textit{pt is a 50yo male with history of CHF who presents with edema.})

This abrupt shift in extractiveness should affect content selection.  In particular, when looking at oracle extractive strategies, we should see clear-cut evidence of \textbf{(1)} 1-2 sentences which are easy to identify as salient (i.e., high lexical overlap with source due to copy-paste), \textbf{(2)} a murkier signal thereafter.  To confirm this, we analyze the sentences selected by the \textsc{Oracle Gain} method, which builds a summary by iteratively maximizing the $R_{12}$ score of the existing summary vis-a-vis the reference.

\begin{figure}[ht]
\begin{center}
\centerline{\includegraphics[width=\columnwidth]{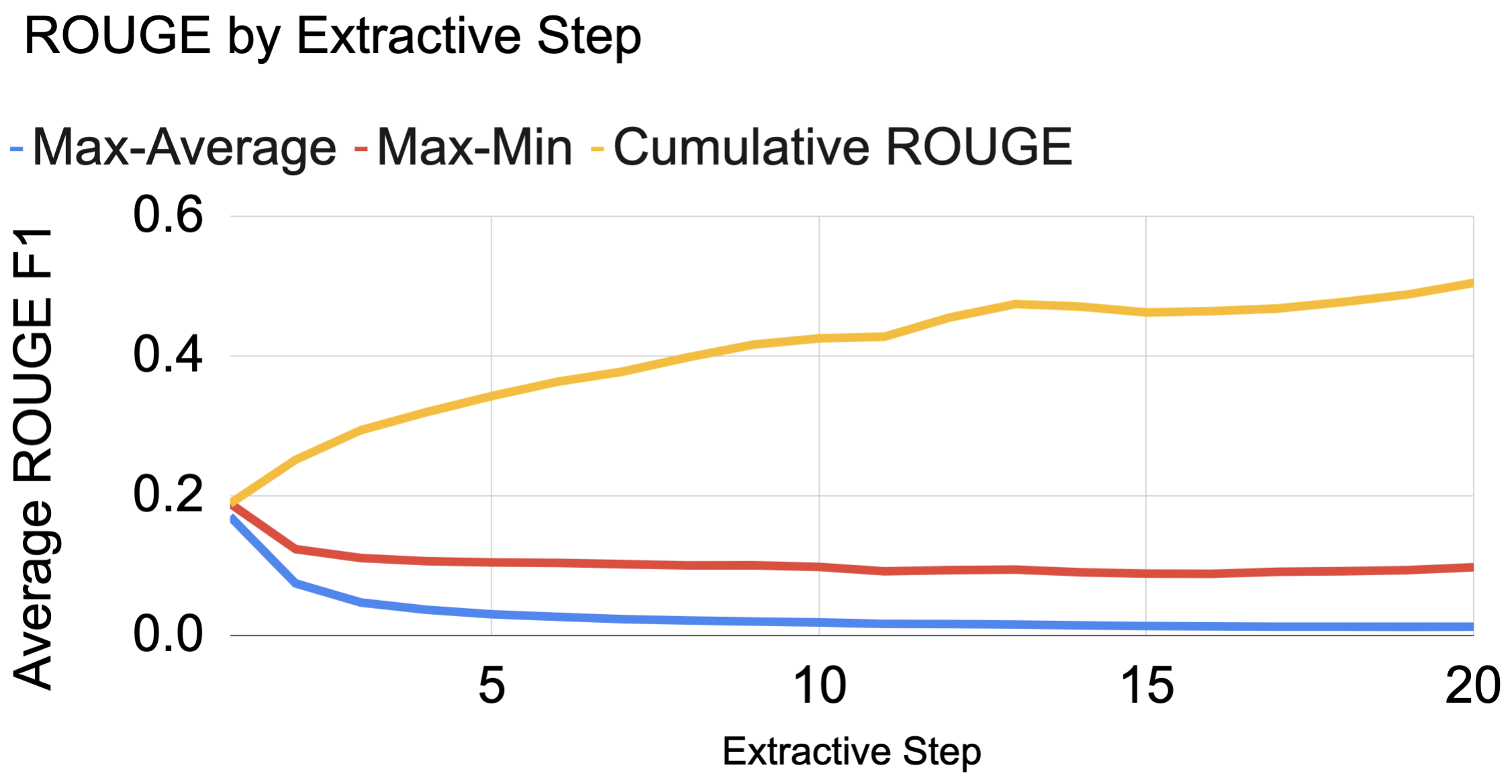}}
\caption{We plot average ROUGE score as summaries are greedily built by adding the sentence with the highest relative ROUGE gain vis-a-vis the current summary, until the gain is no longer positive (\textsc{Oracle Gain}).  We also include the difference between the highest scoring sentence and the average / minimum to demonstrate a weakening sentence selection signal after the top 1-2. }
\label{fig:rouge-by-step}
\end{center}
\vskip -0.2in
\end{figure}

In Figure \ref{fig:rouge-by-step}, two supporting trends emerge.  \textbf{(1)} On average, one sentence accounts for roughly 50\%\footnote{From Table \ref{tab:big-table}, the average $R_{12}$ score is 0.39 for \textsc{Oracle Gain}.  To reconcile this number with respect to Figure \ref{fig:rouge-by-step}, we note that the average oracle summary is far less than the 20 sentence upper bound shown in the chart.} of the overall $R_{12}$ score. \textbf{(2)} Afterwards, the marginal contribution of the next shrinks, as well as the $R_{12}$ gap between the best sentence and the minimum / average, according to the oracle.

There should also be evidence of the copy-paste positional bias impacting content selection.  Table \ref{tab:pos-rank-by-step} reveals that the order in which the \textsc{Oracle Gain} summary is built--by maximal lexical overlap with the partially built summary--roughly corresponds to the true ordering of the summary.  More simply, the summary transitions from extractive to abstractive.

\begin{table}[ht]
    \centering
    \begin{tabular}{c|c}
    \bfseries Extractive & \bfseries Average Rank of Closest \\
    \bfseries Step & \bfseries Reference Sentence \\
    1 & 4.7 \\
    2 & 6.0 \\
    3 & 6.3 \\
    4 & 6.7 \\
    5 & 7.3 \\
    > 5 & 10.1 \\
    \end{tabular}
    \caption{\textsc{Oracle Gain} greedily builds summaries by repeatedly selecting the sentence which maximizes the $R_{12}$ score of the partially built summary.  By linking each extracted sentence to its closest in the reference, we show that this oracle order is very similar to the true ordering of the summary.}
    \label{tab:pos-rank-by-step}
\end{table}

Unsurprisingly, a model (\textsc{ClinNeusum}) trained on \textsc{Oracle Gain} extractions gets progressively worse at mimicking it. Specifically, for each extractive step, there exists a ground-truth ranking of candidate sentences by relative $R_{12}$ gain.  As the relevance gap between source sentences shrinks (from Figure \ref{fig:rouge-by-step}), \textsc{ClinNeusum}'s predictions deviate further from the oracle rank (Table \ref{tab:rank-by-step}).

\begin{table}[ht]
    \centering
    \begin{tabular}{c|c|c}
    \bfseries Extractive & \multicolumn{2}{c}{\bfseries Ground Truth Rank} \\
    \bfseries Step & \bfseries Average & \bfseries Median  \\
    1 & 28 & 7 \\
    2 & 69 & 22 \\
    3 & 74 & 31 \\
    4 & 79 & 39 \\
    5 & 76 & 42 \\
    > 5 & 80 & 60 \\
    \end{tabular}
    \caption{Rank of selected sentence vis-a-vis oracle rank at each extraction step.  A perfectly trained system would have a ground-truth of 1 at each step.}
    \label{tab:rank-by-step}
    \vskip -0.2in
\end{table}

\paragraph{Analysis - Redundancy.} Even though we prevent all baseline methods from generating duplicate sentences (23\% of source sentences have exact match antecedents), there is still a great deal of redundancy in the source notes (i.e., modifications to copy-pasted text).  This causes two issues related to content selection. The first is fairly intuitive - that local sentence extraction propagates severe redundancy from the source notes into the summary and, as a result, produces summaries with low lexical coverage.  We confirm this by examining the performance between the \textsc{Oracle Top-K} and \textsc{Oracle Gain}, which represent summary-unaware and summary-aware variants of the same selection method.  While both extract sentences with the highest $R_{12}$ score, \textsc{Oracle Gain} outperforms because it incorporates redundancy by considering the relative $R_{12}$ gain from an additional sentence.

The second side effect is perhaps more surprising, and divergent from findings in summarization literature.  For most corpora, repetition is indicative of salience.  In fact, methods based on lexical centrality, i.e., TextRank \citep{mihalcea2004textrank} and LexRank \citep{erkan2004lexrank}, still perform very competitively for most datasets. Yet, for \dataset, LexRank barely outperforms a random baseline.  Poor performance is not only due to redundance, but also a weak link between lexical centrality and salience. The Pearson correlation coefficient between a sentence's LexRank score and its $R_{12}$ overlap with the reference is statistically significant ($p=0$) yet weak ($r=0.29$).

Qualitative analysis reveals two principal reasons, both related to copy-paste and/or imported data.  The first relates to the propagation of frequently repeated text which may not be useful for summaries: administrative (names, dates), imported structured data, etc.  The second relates to sentence segmentation. Even though we use a custom sentence splitter, our notes still contain some very long sentences due to imported lists and semi-structured text--a well-documented issue in clinical NLP \citep{leaman2015challenges}. LexRank summaries have a bias toward these long sentences (26.2 tokens versus source average of 10.9), which have a greater chance of containing lexical centroid(s).

\begin{figure}
\centering
\begin{center}
\includegraphics[width=\columnwidth]{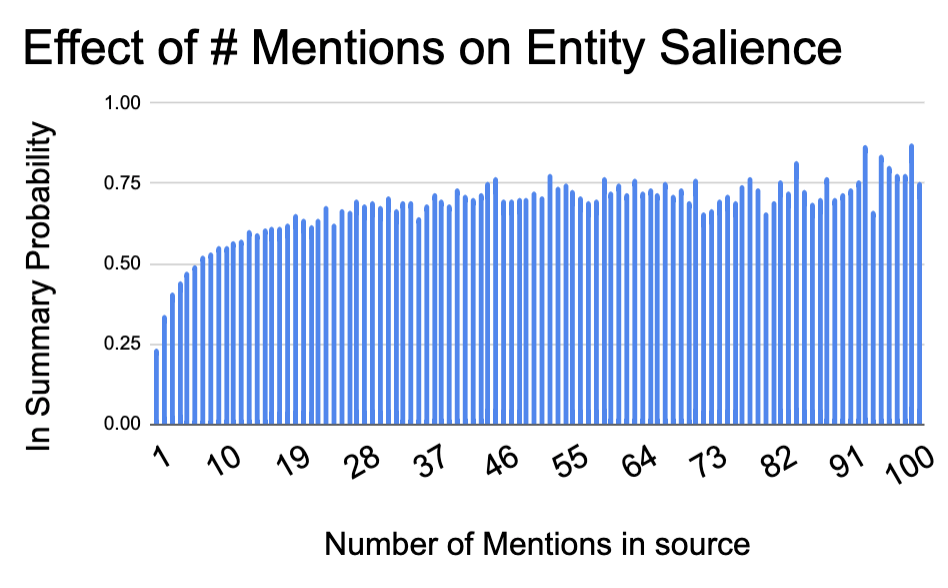}
\caption{Relationship between source entity mentions and probability of inclusion in the summary.
}
\label{fig:source-freq-recall}
\end{center}
\vskip -0.2in
\end{figure}

To bypass some of these issues, however, one can examine the link between centrality and salience at the more granular level of entities. \textbf{Figure \ref{fig:source-freq-recall}} shows a clear-cut positive correlation between source note mention frequency of UMLS concepts and the probability of being included in the summary.

\paragraph{Implications.}

Regarding within-summary variation in \textbf{extractiveness}, we argue for a hybrid approach to balance extraction and abstraction. One of the most widely-used hybrid approaches to generation is the Pointer-Generator (PG) model \citep{see2017get}, an abstractive method which allows for copying (i.e., extraction) of source tokens.  
% \citet{gehrmann2018bottom} enhance the PG model with bottom-up attention to limit the available selection of the PG. 
Another research avenue explicitly decouples the two. These extract-then-abstract approaches come in different flavors: sentence-level re-writing \citep{chen2018fast, bae2019summary}, multi-sentence fusion \citep{lebanoff2019scoring}, and two-step disjoint extractive-abstracive steps \citep{mendes2019jointly}.

While highly effective in many domains, these approaches do not consider systematic differences in extractiveness within a single summary. To incorporate this variance, one could extend the PG model to copy pre-selected long snippets of text.  This would mitigate the problem of copy mechanisms learning to copy very long pieces of text \citep{gehrmann2018bottom} - undesirable for the highly abstractive segments of \dataset.  Span-level extraction is not a new idea \citep{xu2020discourse}, but, to our knowledge, it has not been studied much in otherwise abstractive settings.  For instance, \citet{joshi2020dr} explore patient-doctor conversation summarization and add a penalty to the PG network for over-use of the generator, yet this does not account for intra-summary extractiveness variance.

% As a more drastic decoupling, a traditional IE template-filling \citep{ji2013open} approach might be suitable to generate the standard HPI-like statement at the beginning, before left-to-right generation.

Regarding \textbf{redundancy}, it is clear that, in contrast to some summarization tasks \citep{kedzie2018content}, summary-aware content selection is essential for hospital course summarization.  Given so much noise, massive EHR and cite-specific pre-processing is necessary to better understand the signal between lexical centrality and salience.

\subsection{Summaries are concise yet comprehensive} \label{sec:concise}

\paragraph{tl;dr.} BHC summaries are packed with medical entities, which are well-distributed across the source notes. As such, relations are often not explicit.  Collectively, this difficult task calls for a domain-specific approach to assessing faithfulness.

\paragraph{Analysis - concise}

%The purpose of the BHC is to succinctly convey the major events of the patient's hospital stay to the patient's post-hospital care team. As such, 

We find that summaries are extremely dense with medical entities: $20.9\%$ of summary words are medical UMLS entities, compared to $14.1\%$ in the source notes.  On average, summaries contain $26$ unique entities whereas the source notes contain $265$ --- an entity compression ratio of 10 (versus token-level compression of 43). 

% On average, 73\% of entities in the BHC were found in the source notes while the remaining 27\% did not appear in source text. This could be due to: entities actually conveying new information (e.g., events that happened the day of discharge), hospital documentation ignored by our corpus collection, or faulty named-entity recognition and linking to the UMLS.

\paragraph{Analysis - comprehensive.}

\begin{figure}[!ht]
\centering
\begin{center}
\includegraphics[width=\columnwidth]{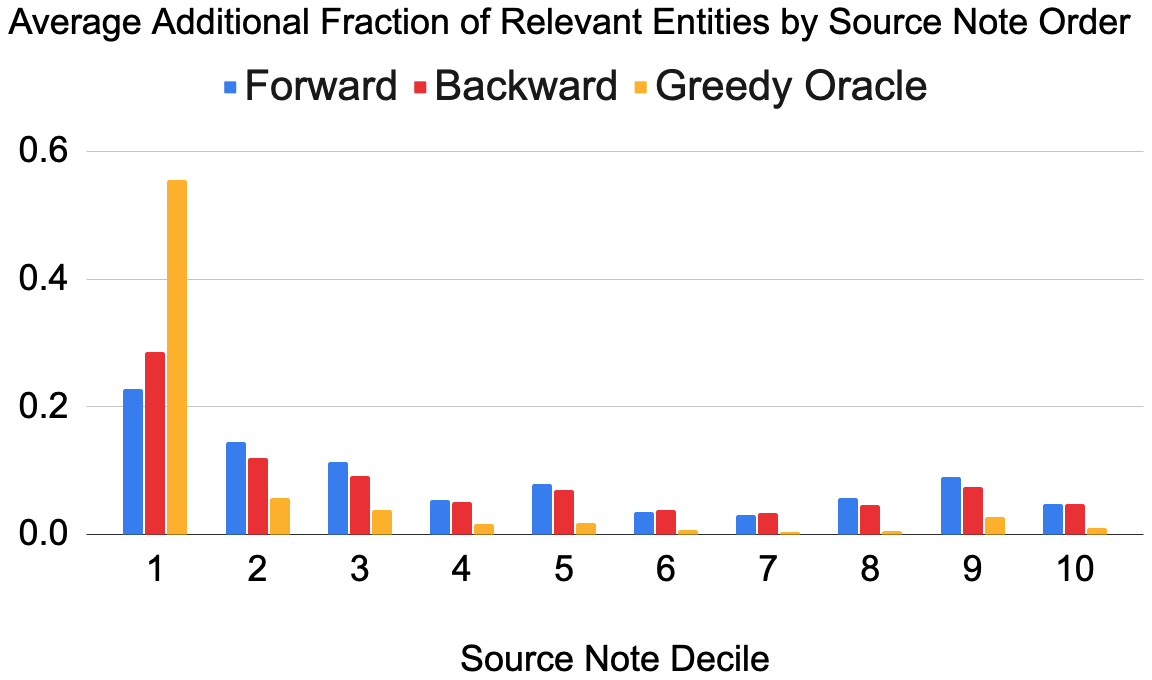}
\caption{The average fraction of additional relevant UMLS entities---present in the summary---from reading a patient's visit notes.  
% Due to large variance in number of source notes per admission, we bucket them into deciles defined by the ordering strategy. 
\textsc{Forward} orders the notes chronologically, \textsc{Backward} the reverse, and \textsc{Greedy Oracle} in order of decreasing entity overlap.}
\label{fig:chronology}
\end{center}
\vskip -0.2in
\end{figure}

\begin{figure}[!ht]
\centering
\begin{center}
\includegraphics[width=\columnwidth]{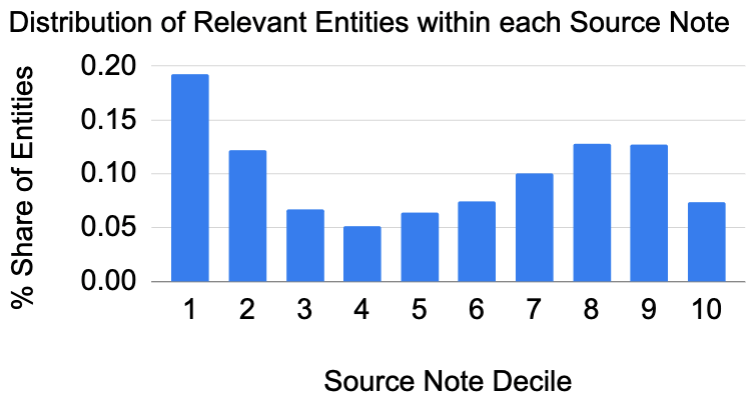}
\caption{The distribution of relevant entities---present in the summary---within an average source note. Source Note Decile refers to the relative position of each mention within a note. Relevant entities appear throughout an average note, with a slight lead bias.}
\label{fig:lead-bias}
\end{center}
\vskip -0.2in
\end{figure}

Many summarization corpora exhibit systematic biases regarding where summary content can be found within source document(s) \citep{dey2020corpora}. On \dataset, we examine the distribution of entities along two dimensions: \textit{macro} considers the differences in entity share across notes, and \textit{micro} considers the differences within each note (i.e., lead bias). \textbf{(1) Macro Ordering.} When looking at the source notes one by one, how much \textit{additional} relevant information (as measured by entities present in the summary) do you get from each new note?  We explore three different orderings: (1) \textsc{Forward} orders the notes chronologically, (2) \textsc{backward} the reverse, and (3) \textsc{Greedy Oracle} examines notes in order of decreasing entity entity overlap with the target.  Given the large variation in number of notes per admission, we normalize by binning notes into deciles. \textbf{Figure \ref{fig:chronology}} shows that it is necessary to read the entire set of notes despite diminishing marginal returns.  One might expect the most recent notes to have the most information, considering present as well as copy-forwarded text. Surprisingly, \textsc{forward} and \textsc{backward} distributions are very similar.  \textsc{Greedy Oracle} gets at the level of information concentration.  On average, the top 10\% of most informative notes cover just over half of the entities found in the summary. We include absolute and percentage counts in Table \ref{tab:ordering}.  \textbf{(2) Micro Ordering.} We plot a normalized histogram of summary entities by relative position within the source documents. Figure \ref{fig:lead-bias} reveals a slight lead bias, followed by an uptick toward the end.  Clinical notes are organized by section: often starting with the past medical history and present illness, and typically ending with the plan for future care.  All are needed to write a complete BHC.

\begin{table}[ht]
    \centering
    \begin{tabular}{l|c|c|c}
    & \multicolumn{2}{c}{Avg Notes to Read} \\
    \bfseries Ordering & \bfseries Number & \bfseries Percent \\
    \textsc{Forward} & 8.5 & 0.80 \\
    \textsc{Backward} & 7.8 & 0.73 \\
    \textsc{Greedy Oracle} & 5.0 & 0.50 \\
    \end{tabular}
    \caption{Number of documents necessary to cover all relevant UMLS entities---present in the summary---according to three different ordering strategies. \textsc{Forward} orders the notes chronologically, \textsc{Backward} the reverse, and \textsc{Greedy Oracle} examines notes in order of decreasing entity overlap with the target.}
    \label{tab:ordering}
    \vskip -0.2in
\end{table}

\paragraph{Implications.}

The fact that entities are so densely packed in summaries makes models more susceptible to factual errors that misrepresent complex relations. On the CNN/DailyMail dataset, \citet{goel2021robustness} reveal performance degradation as a function of the number of entities.  This is magnified for clinical text, where failure to identify which treatments were tolerated or discontinued, or to differentiate conditions of the patient or family member, could lead to serious treatment errors. 

Recently, the summarization community has explored fact-based evaluation.  Yet, many of the proposed methods treat global evaluation as the independent sum of very local assessments.  In the case of QA-based methods, it is a quiz-like aggregation of individual scores to fairly narrow questions that usually seek to uncover the presence or absence of a single entity or relation.  Yet, factoid \citep{chen2018semantic}, cloze-style \citep{eyal2019question, scialom2019answers, deutsch2020towards}, or mask-conditioned question generation \citep{durmus2020feqa} may not be able to directly assess very fine-grained temporal and knowledge-intensive dependencies within a summary.  This is a natural byproduct of the fact that many of the factuality assessments were developed for shorter summarization tasks (i.e., headline generation) in the news domain \citep{cao2018faithful,kryscinski2019neural,maynez2020faithfulness}.  Entailment-based measures to assess faithfulness \citep{pasunuru2018multi, welleck2018dialogue} can capture complex dependencies yet tend to rely heavily on lexical overlap without deep reasoning \citep{falke2019ranking}.

Taken together, we argue for the development of fact-based evaluation metrics which encode a deeper knowledge of clinical concepts and their complex semantic and temporal relations\footnote{\citet{zhang2019optimizing} directly address factuality of clinical text, yet the setting is very different.  They explore radiology report accuracy, which is not a temporal multi-document summarization task.  Additionally, they rely on a smaller IE system tailored specifically for radiology reports \citep{irvin2019chexpert}.}.

\subsection{Summaries have different style and content organization than source notes} \label{misalignment}

\paragraph{tl;dr.} Hospital course summarization involves not only massive compression, but a large style and organization transfer. Source notes are written chronologically yet the way clinicians digest the information, and write the discharge summary, is largely problem-oriented.  With simple oracle analysis, we argue that retrieve-edit frameworks are well-suited for hospital course generation.

\paragraph{Analysis - Style.}  Clinical texts contain many, often obscure, abbreviations \citep{finley2016towards, adams2020zero}, misspellings, and sentence fragments~\citep{demner2009can}. Using a publicly available abbreviation inventory \citep{moon2014sense}, we find that abbreviations are more common in the BHC.  Furthermore, summary sentences are actually longer on average than source sentences (15.8 versus 12.4 words).

\paragraph{Analysis - Organization.}

\begin{figure}[ht]
\begin{center}
\centerline{\includegraphics[width=\columnwidth]{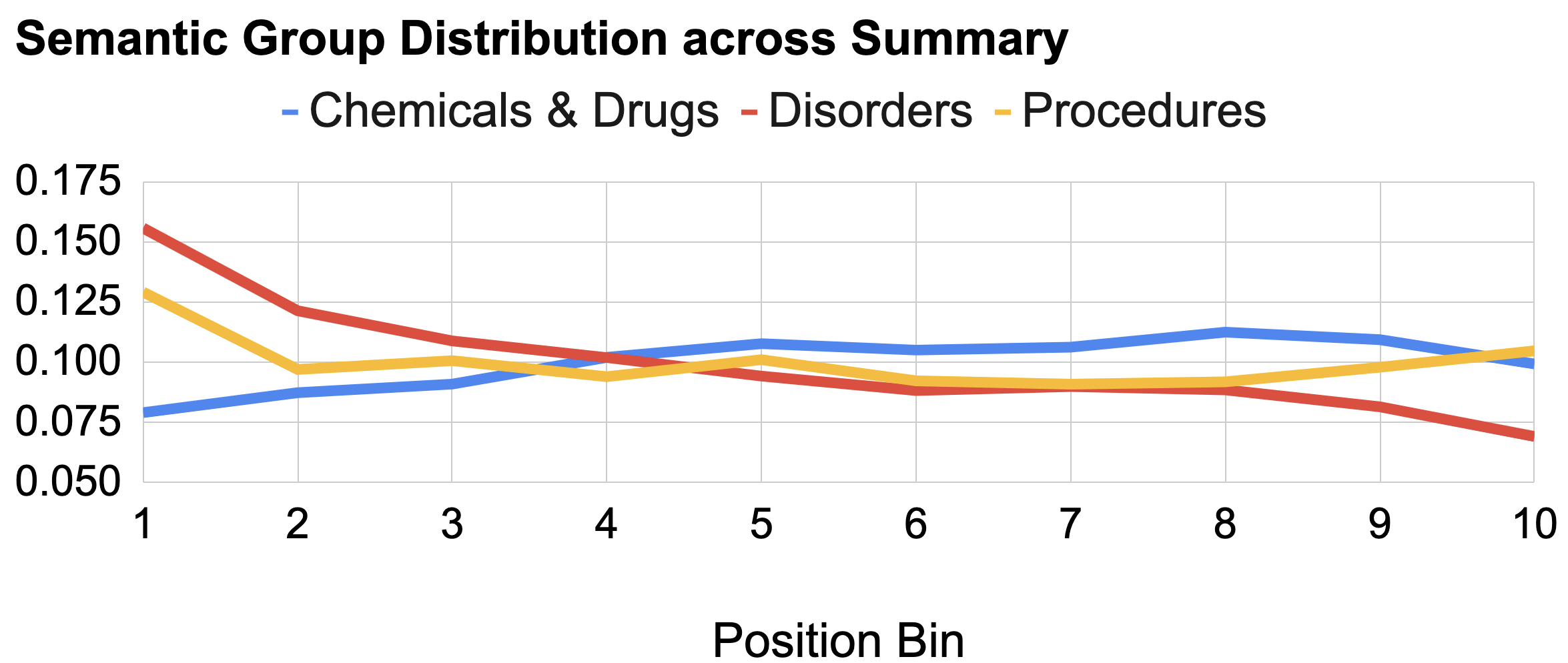}}
\caption{Position of entities within a summary.}
\label{fig:sem-group-positional}
\end{center}
\vskip -0.2in
\end{figure}

\begin{figure}
\centering
\begin{center}
\includegraphics[width=\columnwidth]{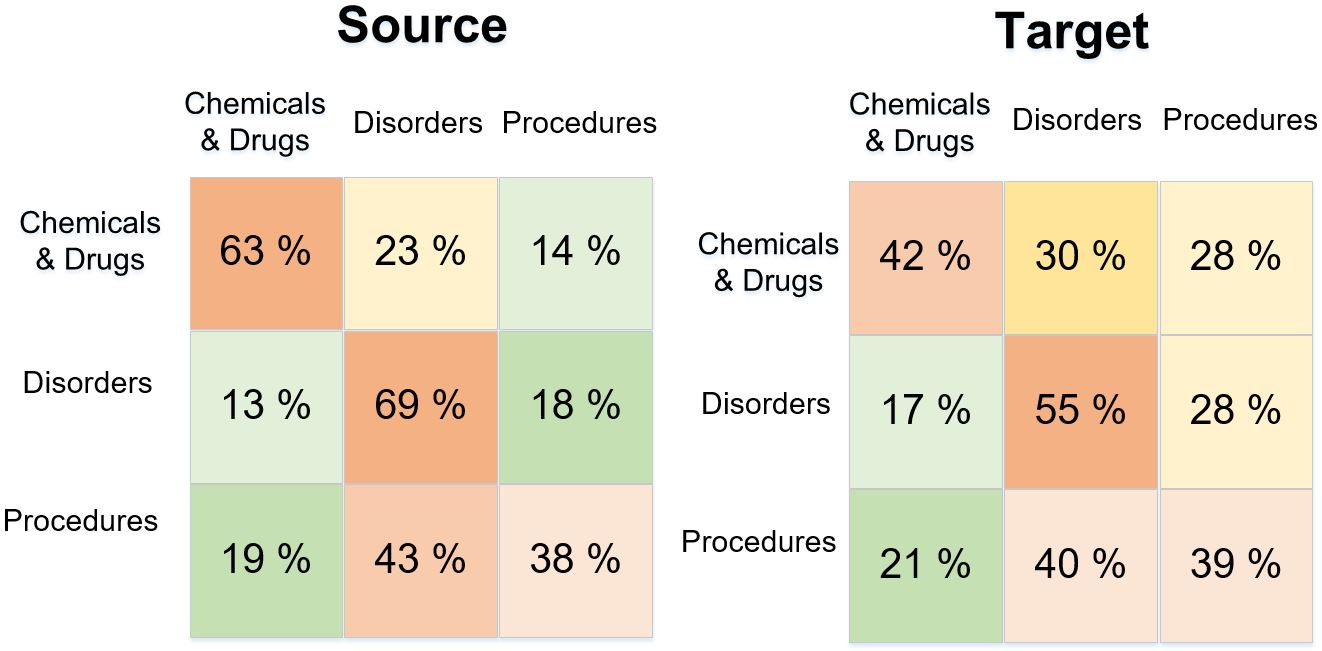}
\caption{Entity Transition Matrices for source notes and target summaries. Summaries have fewer clusters of semantically similar entities,  indicating that entity mentions are woven into a problem-oriented summary.}
\label{fig:ent-trans}
\end{center}
\vskip -0.2in
\end{figure}

Qualitative analysis confirms that most BHCs are written in a problem-oriented fashion \citep{weed1968medical}, i.e., organized around a patient's disorders.  To more robustly analyze content structure, we compare linked UMLS entities at the semantic group level: \textsc{Drugs}, \textsc{Disorders}, and \textsc{Procedures} \citep{mccray2001aggregating}. In particular, we compare \textbf{global} proportions of semantic groups, \textbf{transitions} between entities, as well as \textbf{positional} proportions within summaries.  \textbf{(1) Global.} Procedures are relatively more prevalent in summaries (31\% versus 24\%), maybe because of the emphasis on events happening during the hospitalization. In both summary and source notes, \textsc{Disorders} are the most prevalent (54\% and 46\%, respectively).  Drugs make up 23\% and 22\% of entity mentions in summary and source notes, respectively.  \textbf{(2) Transitions.} From both source and summary text, we extract sequences of entities and record adjacent transitions of their semantic groups in a $3\times3$ matrix. \textbf{Figure \ref{fig:ent-trans}} indicates that summaries have fewer clusters of semantically similar entities (diagonal of the transition matrix).  This transition matrix suggests a problem-oriented approach in which disorders are interleaved with associated medications and lab results.  \textbf{(3) Positional.}  Finally, within summaries, we examine the positional relative distribution of semantic groups and connect it to findings from Section \ref{sec:extractiveness}. In \textbf{Figure \ref{fig:sem-group-positional}}, we first compute the start index of each clinical entity, normalized by the total length, and then group into ten equally sized bins.  The early prevalence of disorders and late prevalence of medications is expected, yet the difference is not dramatic.  This suggests an HPI-like statement up front, followed by a problem oriented narrative.
% As argued in Section \ref{sec:extractiveness}, copy-paste is apropros for the first sentence; after which, problem-oriented models may be necessary.

If there is a material transfer in \textbf{style} and \textbf{content}, we would expect that summaries constructed from other summaries in the dataset would have similar or better lexical coverage than summaries constructed from sentences in the source notes.  To assess this, we compare two oracle baselines, \textsc{sent-align} and \textsc{retrieval}. For each sentence in the summary, we find its closest corollary either in the source text (\textsc{sent-align}) or in other summaries in the dataset (\textsc{retrieval}). While the retrieval method is at a distinct disadvantage because it does not contain patient-specific information and retrieval is performed with BM25 scores, we find both methods yield similar results (\textbf{Table \ref{tab:big-table}}).  An ensemble of \textsc{sent-align} and \textsc{retrieval} performs better than either alone, suggesting that the two types of sources may be complementary.  82\% of this oracle's summary sentences are retrievals. Summaries adapt the style and problem-oriented structure of other summaries, but contain patient-specific information from the source notes.

\begin{figure}[ht]
\begin{center}
\centerline{\includegraphics[width=\columnwidth]{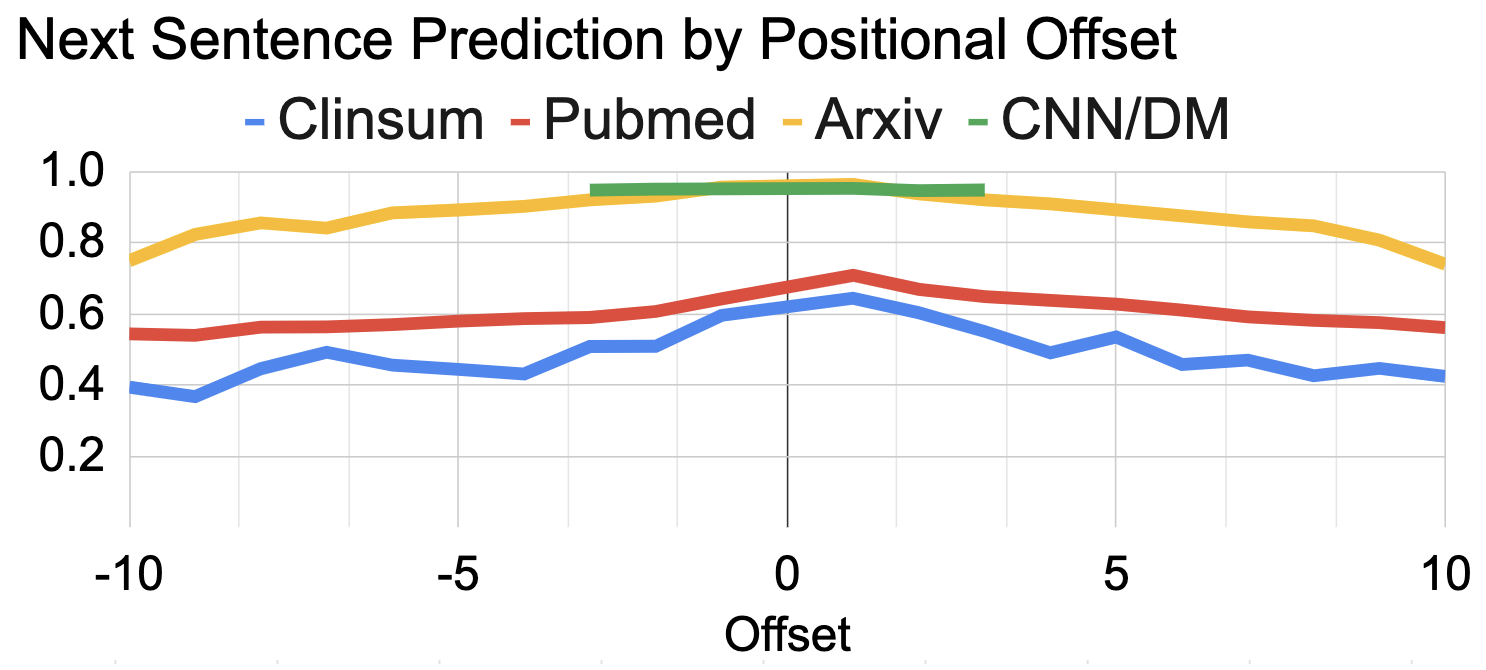}}
\caption{NSP logit by relative position of the next sentence across summaries for several datasets. An offset of 1 corresponds to the true next sentence. }
\label{fig:nsp}
\end{center}
\vskip -0.2in
\end{figure}

\paragraph{Implications.}

Hospital-course summaries weave together disorders, medications, and procedures in a problem-oriented fashion.  It is clear that substantial re-writing and re-organization of source content is needed. One suitable approach is to use the retrieve-rerank-rewrite ($R^3$) framework proposed by \citet{cao2018retrieve}.  To support this notion, more recent work demonstrates that retrieval augmented generation is effective for knowledge-intensive tasks \citep{lewis2020retrieval}, enhances system interpretability \citep{guu2020realm,krishna2020generating}, and can improve LM pre-training \citep{lewis2020pre}\footnote{The related idea of template-based generation has gained traction within the probabilistic community \citep{wiseman2018learning, guu2018generating, wu2019response, he2020learning}.}.  Also, efforts to bridge the gap between template-based and abstractive generation have been successful in the medical domain for image report generation \citep{li2018hybrid}.

% \footnote{The idea of template-based re-writing has also gained traction within the probabilistic community \citep{guu2018generating, wu2019response, he2020learning}.}

In this light, BHC generation could be truly problem-oriented.  The first step would involve selecting salient problems (i.e., disorders) from the source text--a well-defined problem with proven feasibility \citep{van2010corpus}. The second step would involve separately using each problem to retrieve problem-specific sentences from other summaries. These sentences would provide clues to the problem's relevant medications, procedures, and labs. In turn, conceptual overlap could be used to re-rank and select key, problem-specific source sentences.  The extracted sentences would provide the patient-specific facts necessary to re-write the problem-oriented retrieved sentences.

\subsection{Summaries exhibit low lexical cohesion} \label{cohesion}

\paragraph{tl;dr.} Lexical cohesion is sub-optimal for evaluating hospital-course discourse because clinical summaries naturally exhibit frequent, abrupt topic shifts.  Also, low correlation exists between lexical overlap and local coherence metrics.

\paragraph{Analysis.} Entity-based coherence research posits that "texts about the same discourse entity are perceived to be more coherent than texts fraught with abrupt switches from one topic to the next" \citep{barzilay2008modeling}.  Yet, for \dataset summaries, coherence and abrupt topic shifts are not mutually exclusive. An analysis of the entity grids of summaries, presumably coherent, are sparse, with few lexical chains. In fact, over 66\% of the entities in the BHC appear only once.  Of those with multiple mentions, the percentage which appear in adjacent sentences is only 9.6\%.  As in \citet{prabhumoye2020topological}, we also compare coherence with next-sentence prediction (NSP). \textbf{Figure \ref{fig:nsp}} plots the NSP logit by positional offset, where an offset of 1 corresponds to the next sentence, and -1 to the previous. NSP relies on word overlap and topic continuity \citep{bommasani2020intrinsic}, so it makes sense it is lowest for \dataset.

To confirm the hypothesis that ROUGE does not adequately capture content structure, we use the \textit{pairwise ranking} approach to train and evaluate an entity-grid based neural coherence model~\cite{barzilay2008modeling,nguyen2017neural}. 
\textbf{Table \ref{tab:pra}} shows ROUGE and coherence metrics side-by-side for \textsc{Oracle Gain}, which naively orders sentences according to document timestamp, then within-document position, and \textsc{Oracle Sent-Align}, which maintains the structure of the original summary. The poor coherence of \textsc{Oracle Gain} is obscured by comparable ROUGE scores.

\begin{table}[ht]
    \centering
    \begin{tabular}{l|c|c|c}
    \bfseries Summary & \bfseries Acc. & \bfseries R1 & \bfseries R2 \\
    Actual Summary & 0.86 & N/A & N/A \\
    \textsc{Oracle Sent-Align} & 0.75 & 0.52 & 0.30 \\
    \textsc{Oracle Gain} & 0.54 & 0.48 & 0.30 \\
    \end{tabular}
    \caption{Comparison of coherence and ROUGE. Acc. refers to pair-wise ranking accuracy from scoring summaries against random permutations of themselves. }
    % R1 \& R2 F1 entries are repeated from Table \ref{tab:big-table} for illustrative purposes. }
    \label{tab:pra}
    \vskip -0.2in
\end{table}

\paragraph{Implications.}

Content organization is critical and should be explicitly evaluated.  A well-established framework for assessing organization and readability is coherence.  A large strand of work on modeling coherent discourse has focused on topical clusters of entities \citep{azzam1999using, barzilay2002inferring, barzilay2004catching, okazaki2004improving}.  Yet, as shown above, \dataset summaries exhibit abrupt topic shifts and contain very few repeated entities.  The presence and distribution of lexical \citep{morris1991lexical, barzilay1999using} or co-referential \citep{azzam1999using} chains, then, might not be an appropriate proxy for clinical summary coherence.  Rather, we motivate the development of problem-oriented models of coherence, which are associative in nature, and reflect a deeper knowledge about the relationship between disorders, medications, and procedures. The impetus for task-tailored evaluation metrics is supported by recent meta analyses \citep{fabbri2020summeval, bhandari2020re}.

\subsection{BHC summaries are silver-standard} \label{silver-standard}
\paragraph{tl;dr.}  Discharge summaries and their associated BHC sections are frequently missing critical information or contain excessive or erroneous content.  Modeling efforts should address sample quality.

\paragraph{Analysis.}
\citet{kripalani2007deficits} find that discharge summaries often lack important information including diagnostic test results (33-63\% missing) treatment or hospital course (7-22\%), discharge medications (2-40\%), test results pending at discharge (65\%), patient/family counseling (90-92\%), and follow-up plans (2-43\%). The quality of the reporting decreases as the length of the discharge summary increases, likely due to copy-pasted information~\citep{van1999necessary}.

These quality issues occur for a number of reasons: (1) limited EHR search functionality makes it difficult for clinicians to navigate through abundant patient data~\citep{christensen2008instant}; (2) multiple clinicians contribute to incrementally documenting care throughout the patient's stay; (3) despite existing guidance for residents, clinicians receive little to no formal instruction in summarizing patient information~\citep{ming2019discharge}; and (4) clinicians have little time for documenting care.

\paragraph{Implications.}

Noisy references can harm model performance, yet there is a rich body of literature to show that simple heuristics can identify good references \citep{bommasani2020intrinsic} and/or filter noisy training samples \citep{rush2015neural, akama2020filtering, matsumaru2020improving}. Similar strategies may be necessary for hospital-course generation with silver-standard data.  Another direction is scalable reference-free evaluations \citep{shafieibavani2018summarization, narayan2019highres, sellam2020bleurt, gao2020supert, vasilyev2020fill}.

% It would be interesting to see if the same metrics are as equally discriminative for clinical summaries.  For example, BHC sections that are unexpectedly short or long compared to other patients with similar disease complexity may be missing information or contain excessive copy-pasted text. While many summaries are problem-oriented, less experienced clinicians tend to write chronological summaries that provide a too-detailed laundry list of all events.

\section{Conclusion}

Based on a comprehensive analysis of clinical notes, we identify a set of implications for hospital-course summarization on future research.  For modeling, we motivate \textbf{(1)} the need for dynamic hybrid extraction-abstraction strategies (\ref{sec:extractiveness}); \textbf{(2)} retrieval-augmented generation (\ref{misalignment}); and \textbf{(3)} the development of heuristics to assess reference quality (\ref{silver-standard}).  For evaluation, we argue for \textbf{(1)} methods to assess factuality and discourse which are associative in nature, i.e., incorporate the complex inter-dependence of problems, medications, and labs (\ref{sec:concise}, \ref{cohesion}); and \textbf{(2)} scalable reference-free metrics (\ref{silver-standard}).

% Each takeaway is a standalone research direction with its \textbf{tl;dr} an abstract blueprint.

\section{Ethical Considerations}
\xhdr{Dataset creation}
Our \dataset dataset contains protected health information about patients. We have received IRB approval through our institution to access this data in a HIPAA-certified, secure environment. To protect patient privacy, we cannot release our dataset, but instead describe generalizable insights that we believe can benefit the general summarization community as well as other groups working with EHR data.

\xhdr{Intended Use \& Failure Modes}
The ultimate goal of this work is to produce a summarizer that can generate a summary of a hospital course, and thus support clinicians in this cognitively difficult and time-consuming task. While this work is a preface to designing such a tool, and significant advances will be  needed to achieve the robustness required for deployment in a clinical environment, it is important to consider the ramifications of this technology at this stage of development. We can learn from existing clinical summarization deployed~\cite{pivovarov2016can} and other data-driven clinical decision support tools~\cite{chen2020ethical}. As with many NLP datasets, \dataset likely contains biases, which may be perpetuated by its use. 
%For example, if the model is trained on female patients whose heart attack symptoms are undiagnosed, the models can learn to perpetuate this underdiagnosis. 
There are a number of experiments we plan to carry out to identify documentation biases and their impact on summarization according to a number of dimensions such as demographics (e.g., racial and gender), social determinents of health (e.g., homeless individuals), and clinical biases (e.g., patients with rare diseases). %The data contains fewer admissions for underrepresented populations and diseases and therefore may perform worse on racial minorities, transgender individuals, and patients with rare diseases. 
Furthermore, deployment of an automatic summarizer may lead to automation bias~\citep{goddard2012automation}, in which clinicians over rely on the automated system, despite controls measures or verification steps that might be built into a deployed system. Finally, medical practices and EHRs systems constantly change, and this distribution drift can cause models to fail if they are not updated. As the NLP community continues to develop NLP applications in safety-critical domains, we must carefully study how can can build robustness, fairness, and trust into these systems. 

\section*{Acknowledgements}

We thank Alex Fabbri and the NAACL reviewers for their constructive, thoughtful feedback. This work was supported by NIGMS award R01 GM114355 and NCATS award U01 TR002062.

\bibliography{anthology,main_fixed}
\bibliographystyle{acl_natbib}

\appendix

\begin{table*}[t]
    \centering
    \small
    \begin{tabular}{c|l|c|c|c|c|c|c|c|c}
    & & & Comp. & \multicolumn{2}{c|}{Extractiveness} & \multicolumn{2}{c|}{Summary} & Source \\
     & \bfseries Dataset & \# Docs & Ratio & Coverage & Density & \# words & \# sents & \# words \\
    \hline
    \multirow{7}{*}{SDS}
    & Gigaword \citep{Rush_2015} & 4mn & 3.8 & 0.58 & 1.1 & 8.3 & 1 & 31.4 \\
    % & NYT & 654k & 12.0 & & & 44.9 & 2.0 & 795.9 \\
    & CNN/DM \citep{nallapati-etal-2016-abstractive} & 312k & 13.0 & 0.80 & 3.0 & 55.6 & 3.8 & 789.9 \\
    & Newsroom \citep{grusky2018newsroom} & 1.2mn & 43.0 & 0.82 & 9.6 & 30.4 & 1.4 & 750.9 \\
    & XSum \citep{narayan2018don} & 226k & 18.8 & 0.57 & 0.89 & 23.3 & 1.0 & 431.1 \\
    & Arxiv \citep{cohan2018discourse} & 215k & 39.8 & 0.92 & 3.7 & 292.8 & 9.6 & 6,913.8 \\
    & PubMed \citep{cohan2018discourse} & 133k & 16.2 & 0.90 & 5.9 & 214.4 & 6.9 & 3,224.4 \\
    & BigPatent \citep{sharma2019bigpatent} & 1.3mn & 36.4 & 0.86 & 2.4 & 116.5 & 3.5 & 3,572.8 \\
    \hline
    \multirow{5}{*}{MDS}
    & WikiSum \citep{liu2018generating} & 2.3mn & 264.0 & N/A & N/A & 139.4 & N/A & 36,802.5 \\
    & Multi-News \citep{fabbri2019multi} & 56k & 8.0 & 0.68 & 3.0 & 263.7 & 10 & 2,103.5 \\
    & SOAP \citep{krishna2020generating} & 7k & 4.7 & N/A & N/A & 320 & N/A & 1,500 \\
    % & GameWikiSum \citep{antognini2020gamewikisum} & 147k & 24.4 & 321 & N/A & 7,815 \\
    & \dataset (ours) & 110k & 45.2 & 0.83 & 13.1 & 261.9 & 17.7 & 11,838.7 \\
    \end{tabular}
    \caption{Basic statistics for single-document (SDS) and multi-document (MDS) summarization datasets. For multi-document summarization (MDS), \# Source words are aggregated across documents. Compression ratio is the average ratio of source words to summary words.  Extractiveness metrics (coverage and density) come from \citet{grusky2018newsroom} and, for consistency, are calculated using the official \href{https://github.com/lil-lab/newsroom}{code} across the validation set for each dataset.  Spacy tokenization is performed before extracting fragments.  Other corpus statistics are pulled from either the corresponding paper or Table 1 in \citet{sharma2019bigpatent}.  Entries are filled with N/A because the dataset is private \citep{krishna2020generating}, or too expensive to generate \citep{liu2018generating}. The Gigaword SDS dataset comes from the annotated Gigaword dataset \citep{graff2003english, napoles2012annotated}}
    \label{tab:corpora}.
\end{table*}

\section{Additional Dataset Description} \label{app:sum-datasets}

Based on advice from clinicians, we rely on the following subset of notes as source documents: ``Admission notes'', which convey the past medical history of a patient, ongoing medications, and a detailed description of chief complaint; ``Progress notes'', which convey a daily report about patient status and care as well as to-do lists for next day; and ``Consult notes'', which document specialist consultations. The dataset does not contain any structured data, documentation from past encounters, or other note types (e.g., nursing notes, social work, radiology reports)~\citep{reichert2010cognitive}.\footnote{We note that most structured data fields are now automatically imported into input notes, and, similarly, findings of reports are available in notes.}  Additionally, we remove all visits without at least one source note and at least one Brief Hospital Course target section and exclude notes with less than 25 characters. For computational and modeling feasibility, we bound the minimum and maximum lengths for the source and target texts. We exclude visits where the source notes are collectively over $20,000$ tokens ($<10\%$ of visits) or are shorter than the Brief Hospital Course.  Finally, we exclude visits where the Brief Hospital Course section is less than 25 characters and greater than 500 tokens to remove any incorrectly parsed BHC sections.

\section{Local Coherence Model Details} \label{app:coherence}

The underlying premise of the entity-grid model is that ``the distribution of entities in locally coherent texts exhibits certain regularities" \citep{barzilay2008modeling}.  The paper defines entities as coreferent noun phrases, while we use UMLS entities.  Additionally, \citet{barzilay2008modeling} add syntactic role information to the grid entries, whereas, without reliable parses, we denote a binary indicator of entity presence.  As is common practice, we learn to rank the entity grid of a text more highly than the same entity grid whose rows (sentences) have been randomly shuffled. Inspired by \citet{mohiuddin2018coherence}, we first project the entity grid entries onto a shared embedding space whose vocabulary consists of all the UMLS CUIs and a special <empty> token.  As in \citet{nguyen2017neural}, we then learn features of original and permuted embedded grids by separately applying 1-D convolutions.  Finally, scalars produced by the siamese convolutional networks are used for pairwise ranking.

\section{\textsc{ClinNeusum} Details} \label{clin-neusum-details}

As in \citet{nallapati2016summarunner} and \citet{zhou2018neural}, we extract ground truth extraction labels by greedily selecting sentences which maximize the relative ROUGE gain ($R_{12}$) from adding an additional sentence to an existing summary. We use basic heuristics to scale the model efficiently to a dataset with such a large set of candidate sentences. To avoid inclusion of sentences with spurious, small relevance based on ROUGE, we filter out extractive steps with a weak learning signal - extractive steps for which either the ROUGE improvement of the highest scoring sentence is less than 1\%, or the differential between the least and most relevant sentences is less than 2\%.  Furthermore, based on manual evaluation, we take steps to reduce the size of the candidate sentence set provided to the model sees during training. First, we de-duplicate sentences and remove sentences with no alphabetical letters or a token count less than 3.  Then, we randomly remove source sentences without any lexical overlap with the summary with a probability determined by source length.  This produces a train-test bias, but it is minor because most of the removed sentences are consistently irrelevant (i.e., dates, numerical lists, signature lines, etc.).  During training, we randomly sample a single extractive step whose objective is to maximize the KL-Divergence between the model-generated score distribution over sentences and a temperature-smoothed softmax over the relative ROUGE gain.

Similarly to the Neusum model, we employ a simple LSTM-based, hierarchical architecture. We project source and target words onto a shared embedding space.  Then, separately, we pass word embeddings to a bi-LSTM sentence encoder.  We use the concatenated hidden states from the forward and backward pass as input to another bi-LSTM document encoder.  We ignore document boundaries in this setup.  We treat the concatenation of the sentence-level hidden state from the sentence encoder and the corresponding hidden state from the document encoder as the final sentence-level representation.  Then for each candidate source sentence, we attend to each sentence in the existing summary to compute a summary-aware sentence-representation.  Finally, we concatenate both representations and pass through three fully connected layers with Tanh activation.  The output is a single scalar score for which we compute the softmax over all candidate sentences.  We compare to this distribution to the empirical relative ROUGE distribution\footnote{As in the Neusum model, we first min-max normalize the raw ROUGE gains, and then apply a temperature scalar of 5 before computing the softmax.}.  For inference, we greedily extract sentences until the target of 13 sentences (validation average) is reached.

\section{A Note on Copy-Paste in Clinical Text} \label{app:note-on-copypaste}

Researchers have explored unintended side effects of copy-paste along many different dimensions: information bloat, reporting errors and incoherence from outdated or inconsistent information \citep{hirschtick2006copy, yackel2006copy, siegler2009copy, o2009physicians, tsou2017safe}, and quantifying redundancy \citep{wrenn2010quantifying,zhang2011evaluating, cohen2013redundancy}. Quantifying redundancy is non-trivial because copy-paste occurs at different granularities and, quite often, the pasted text is modified. We do not seek to replicate these studies on \dataset. Rather, we examine the impact on summary \textbf{extractiveness} and \textbf{redundancy}.

\end{document}